\pdfoutput=1

\documentclass[11pt]{article}

\usepackage{emnlp2021}

\usepackage{times}
\usepackage{latexsym}

\usepackage[T1]{fontenc}

\usepackage[utf8]{inputenc}

\usepackage{microtype}

\title{
Conversational Multi-Hop Reasoning with \\
Neural Commonsense Knowledge and Symbolic Logic Rules
}
\makeatletter
\newcommand{\printfnsymbol}[1]{%
  \textsuperscript{\@fnsymbol{#1}}%
}
\author{Forough Arabshahi\thanks{~~Equal contribution} \\
  Facebook \\
  \texttt{forough@fb.com} \\\And
  Jennifer Lee\footnotemark[1] \\
  Facebook \\
  \texttt{
jenniferlee98@fb.com} \\ \And
  Antoine Bosselut \\
  EPFL \\
  \texttt{
antoine.bosselut@epfl.ch}\\ \AND Yejin Choi \\
  University of Washington \\
  \texttt{yejin@cs.washington.edu} \\ \And 
  Tom Mitchell \\
  Carnegie Mellon University\\
  \texttt{ tom.mitchell@cmu.edu}} 

\usepackage{xcolor, soul}
\usepackage{amsmath, amsfonts, dsfont, mathtools}
\usepackage{txfonts}
\usepackage{bbold}
\usepackage{makecell}

\usepackage{graphicx}
\usepackage{subcaption}
\usepackage{tabularx}
\usepackage{booktabs}
\usepackage{listings}
\usepackage[switch]{lineno}
\usepackage{multirow}
\usepackage{diagbox}

\definecolor{bondiblue}{rgb}{0.0, 0.58, 0.71}
\definecolor{antiquefuchsia}{rgb}{0.57, 0.36, 0.51}
\definecolor{orange}{rgb}{1,0.5,0}
\definecolor{brickred}{rgb}{0.8, 0.25, 0.33}
	\definecolor{gray(x11gray)}{rgb}{0.75, 0.75, 0.75}
\definecolor{lavendergray}{rgb}{0.77, 0.76, 0.82}
\definecolor{lightgray}{rgb}{0.83, 0.83, 0.83}
\definecolor{snow}{rgb}{1.0, 0.98, 0.98}
\definecolor{splashedwhite}{rgb}{1.0, 0.99, 1.0}
\definecolor{timberwolf}{rgb}{0.86, 0.84, 0.82}
\definecolor{seashell}{rgb}{1.0, 0.96, 0.93}
\definecolor{whitesmoke}{rgb}{0.96, 0.96, 0.96}
\definecolor{platinum}{rgb}{0.9, 0.89, 0.89}
\definecolor{pearl}{rgb}{0.94, 0.92, 0.84}
\definecolor{palepink}{rgb}{0.98, 0.85, 0.87}
\definecolor{oldlace}{rgb}{0.99, 0.96, 0.9}
\definecolor{mistyrose}{rgb}{1.0, 0.89, 0.88}
\definecolor{magnolia}{rgb}{0.97, 0.96, 1.0}
\definecolor{lavenderblush}{rgb}{1.0, 0.94, 0.96}
\definecolor{atomictangerine}{rgb}{1.0, 0.6, 0.4}
\definecolor{babyblue}{rgb}{0.54, 0.81, 0.94}
\definecolor{celadon}{rgb}{0.67, 0.88, 0.69}
\definecolor{darkpastelpurple}{rgb}{0.59, 0.44, 0.84}
\definecolor{flamingopink}{rgb}{0.99, 0.56, 0.67}
\definecolor{bluebell}{rgb}{0.64, 0.64, 0.82}
\definecolor{lavenderblue}{rgb}{0.8, 0.8, 1.0}
\definecolor{amethyst}{rgb}{0.6, 0.4, 0.8}
\definecolor{ao(english)}{rgb}{0.0, 0.5, 0.0}
\definecolor{cadmiumorange}{rgb}{0.93, 0.53, 0.18}
\definecolor{darkorange}{rgb}{1.0, 0.55, 0.0}
\definecolor{flame}{rgb}{0.89, 0.35, 0.13}
\definecolor{internationalorange}{rgb}{1.0, 0.31, 0.0}
\definecolor{alizarin}{rgb}{0.82, 0.1, 0.26}
\definecolor{cadmiumred}{rgb}{0.89, 0.0, 0.13}
\sethlcolor{lavenderblush}
\definecolor{candyapplered}{rgb}{1.0, 0.03, 0.0}
\definecolor{carminered}{rgb}{1.0, 0.0, 0.22}
\definecolor{carminepink}{rgb}{0.92, 0.3, 0.26}
\definecolor{coralred}{rgb}{1.0, 0.25, 0.25}
\definecolor{denim}{rgb}{0.08, 0.38, 0.74}
\definecolor{blue-violet}{rgb}{0.54, 0.17, 0.89}

\newcommand{\coolname}{$\varmathbb{CLUE}$}
\newcommand{\textGoal}{{\fontfamily{lmtt}\selectfont goal\,}}
\newcommand{\textAction}{{\fontfamily{lmtt}\selectfont action\,}}
\newcommand{\textState}{{\fontfamily{lmtt}\selectfont state\,}}

\newcommand{\blueTemplate}[2][babyblue]{ {\sethlcolor{#1} \hl{#2}} }

\newcommand{\purpleTemplate}[2][lavenderblue]{ {\sethlcolor{#1} \hl{#2}} }

\newcommand{\blank}{\textcolor{purple}{$(\,\cdot\,)$}}

\newcommand{\comet}{$\mathbb{COMET}$}
\newcommand\cometemoji{\raisebox{-2pt}{\includegraphics[width=0.9em]{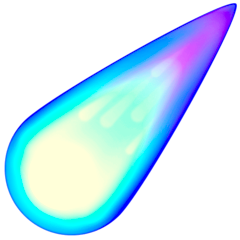}}}

\newcommand{\prologTerm}[1]{{\fontfamily{lmss}\selectfont \hl{#1}}}

\lstdefinelanguage{corgi}{
sensitive=false,
alsoletter={.},
moredelim=[is][\color{blue-violet}]{<}{>},
moredelim=[is][\color{denim}]{[}{]},
moredelim=[is][\color{internationalorange}]{``}{''},
keywords=[10]{...,user,statement,command,megaCORGI,one.,two.,three.,four.,five.,six.,>},
keywordstyle=[10]{\bf},
keywords=[1]{>},
keywordstyle=[1]{\color{alizarin}},
}

\begin{document}

\maketitle

\begin{abstract}
    One of the challenges faced by conversational agents is their inability to identify unstated \emph{presumptions} of their users' commands, a task trivial for humans due to their common sense. 
    In this paper, we propose a zero-shot commonsense reasoning system for conversational agents in an attempt to achieve this. Our reasoner uncovers unstated presumptions from user commands satisfying a general template of {\bf if-(\textState), then-(\textAction), because-(\textGoal)}. 
    Our reasoner uses a state-of-the-art transformer-based generative commonsense knowledge base (KB) as its source of background knowledge for reasoning. 
    We propose a novel and iterative knowledge query mechanism to extract multi-hop reasoning chains from the neural KB which uses symbolic logic rules to significantly reduce the search space. 
    Similar to any KBs gathered to date, our commonsense KB is prone to missing knowledge. 
    Therefore, we propose to conversationally elicit the missing knowledge from human users with our novel dynamic question generation strategy, which generates and presents contextualized queries to human users. 
    We evaluate the model with a user study with human users that achieves a 35\% higher success rate compared to SOTA.
\end{abstract}

\section{Introduction}
Conversational agents are becoming prominent in our daily lives thanks to advances in speech recognition, natural language processing and machine learning. 
However, most conversational agents still lack commonsense reasoning, preventing them from engaging in rich conversations with humans.

Recently, \citet{corgi} proposed a commonsense reasoning benchmark task for conversational agents that contains natural language commands given to an agent by humans. 
These commands follow a general template of: {\bf ``If (\textState~holds), Then (perform \textAction), Because (I want to achieve \textGoal)''}. 
We refer to commands satisfying this template as if-then-because commands. As stated in \citet{corgi}, humans often under-specify conditions on the if-portion (\,\textState) and/or then-portion (\,\textAction) of their commands. These under-specified conditions are referred to as \emph{commonsense presumptions}. For example, consider the command, ``If it's going to rain in the afternoon \blank~Then remind me to bring an umbrella \blank~Because I want to remain dry'', where \blank~indicates the position of the unstated commonsense presumptions. The presumptions for this command are \textcolor{purple}{(}and I am outside\textcolor{purple}{)} and \textcolor{purple}{(}before I leave the house\textcolor{purple}{)}, respectively. The goal in this task is to infer such commonsense presumptions given if-then-because commands.

\begin{figure}[t]
    \centering
    \includegraphics[width=\columnwidth]{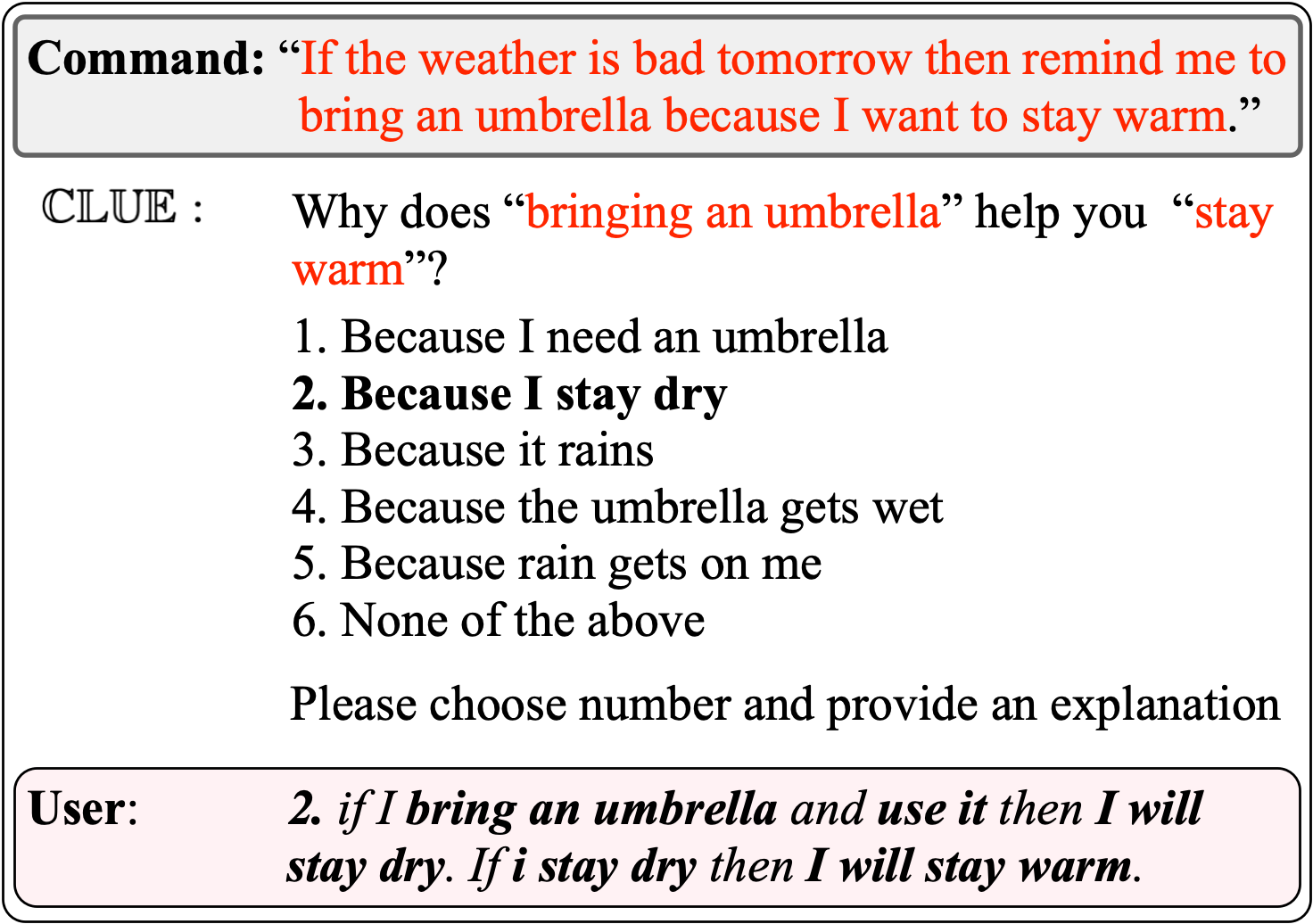}
    \caption{\coolname's conversation with a human. 
    The multiple choices {\bf 1} through {\bf 5} are commonsense knowledge obtained from \comet~using our multi-hop reasoner, ranked by the reasoner's confidence. Since the user chooses \textbf{Option 2} (Because I stay dry), \coolname~selects this reasoning path as the final reasoning chain.}
    \label{fig:dialog}
\end{figure}
\begin{figure*}[t]
    \centering
    \includegraphics[width=\textwidth]{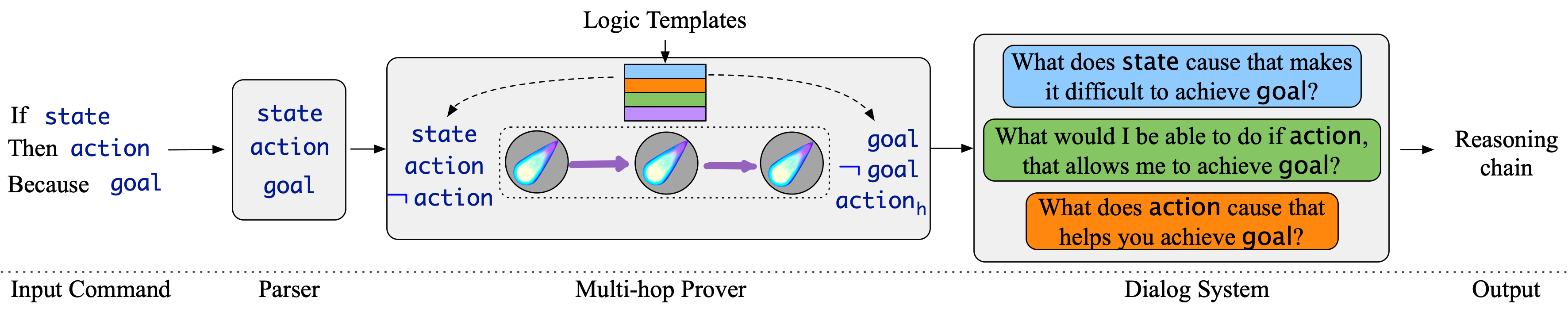}
\caption{\coolname~diagram. \coolname~has three main components: The parser, the multi-hop reasoner (prover) and the dialog system. Given an if-then-because command, the parser extracts independent natural language clauses for the \textState, \textAction and \textGoal. The prover extracts multi-hop reasoning chains given the logic templates using our neural commonsense KB, \comet~that indicates how the \textAction achieves the \textGoal when the \textState holds. The extracted reasoning chains go through the dialog system that generates template-dependent questions and converses with a human who either validates the returned proofs or contributes novel commonsense knowledge if the proofs are incorrect.}
    \label{fig:CLUE_diagram}
\end{figure*}

In this paper, we propose the ConversationaL mUlti-hop rEasoner (\coolname)
for this task which performs zero-shot reasoning. \coolname~extracts a multi-hop reasoning chain that indicates how the \textAction~leads to the \textGoal~when the \textState~holds. For example, the simplified reasoning chain for the previous example is, ``if I am reminded to bring an umbrella before I leave the house, then I have an umbrella'', ``if I have an umbrella and it rains when I am outside, then I can use the umbrella to block the rain'', ``if I block the rain, then I remain dry''. Additional commonsense knowledge provided by the reasoning chain is considered a commonsense presumption of the input command. In order to construct the multi-hop reasoning chain, we develop a novel reasoning system that uses a few symbolic logic templates
to prune the exponential reasoning search space, resulting in significantly improved generated commonsense knowledge. 

Our multi-hop reasoner uses a state-of-the-art (SOTA) transformer-based commonsense knowledge model called COMmonsEnse Transformers (\comet\cometemoji) \cite{Bosselut2019COMETCT} as its source of background knowledge and is the first time reasoning task is tackled using a large scale KB. Knowledge models can be used in place of KBs, but they are more flexible in terms of information access, so we use this term interchangeably with KB. 

Despite being a SOTA KB, \comet{} still misses requisite knowledge for reasoning about if-then-because commands. In fact, many if-then-because commands often fall under the long tail of tasks for which there is too little knowledge available in any commonsense KBs. For example, one of the commands in the dataset is: ``If I get an email with subject the gas kiln fired, then send me a list of all the pots I put on the glaze shelf between the last firing and now, because I want to pick up the pots from the studio.'' 
It is unlikely for any knowledge source to contain a fact that is contextually relevant to this command. More importantly, it is also unlikely that a command requiring the same type of reasoning will occur again in the future, making it cost-ineffective to manually annotate.

To overcome this, we propose conversationally eliciting missing knowledge from human users. 
Conversation with users is readily available since \coolname~is developed for conversational agents. We develop a novel question generation strategy that uses \comet~to generate questions in the context of the input if-then-because commands, allowing \coolname~to acquire contextual feedback from humans. We evaluate our reasoner by conducting a user study with humans. 
\paragraph{Summary of Results and Contributions:} 
The contributions of this paper are two-fold. First, we propose a novel conversational commonsense reasoning approach called ConversationaL mUlti-hop rEasoner (\coolname) that incorporates Commonsense Transformers (\comet), a large scale neural commonsense KB, as its main source of background knowledge. 
We propose a multi-hop knowledge query mechanism that extracts reasoning chains by iteratively querying \comet. This mechanism uses, for the first time, a few symbolic logic templates to prune the reasoning search space, significantly improving the quality of the generated commonsense knowledge. 
Second, we propose a conversational knowledge acquisition strategy that uses the knowledge extracted from \comet~to dynamically ask contextual questions to humans whenever there is missing knowledge. 
We ran a user study and evaluated our proposed approach using real human users. Our results show that \coolname~achieves 
a 35\% higher success rate compared to a baseline, which uses a less sophisticated conversational interface and a smaller background knowledge on the benchmark dataset. We also extensively evaluate the performance of the reasoner in an isolated non-conversational setting and empirically re-iterate the need for conversational interactions. \coolname's components were not trained on the benchmark dataset. 
Therefore, our results assess the performance of \coolname's zero-shot reasoning.
\section{Background and Notation}
\label{sec:background}
In this section, we briefly re-introduce the benchmark task, the logic templates, and \citeauthor{corgi}'s reasoning engine CORGI (COmmonsense ReasoninG By Instruction), along with an overview of \comet~\cite{Bosselut2019COMETCT}. 
\subsection{If-Then-Because Commands}
The benchmark dataset \citep{corgi} contains natural language commands given to conversational agents. These commands follow the template {\bf ``If-(\textState), Then-(\textAction), Because-(\textGoal)''}. 
The if-clause is referred to as the \textState, the then-clause as the \textAction~and the because-clause as the \textGoal.

\begin{figure}[t]
    \centering
    \includegraphics[width=\columnwidth]{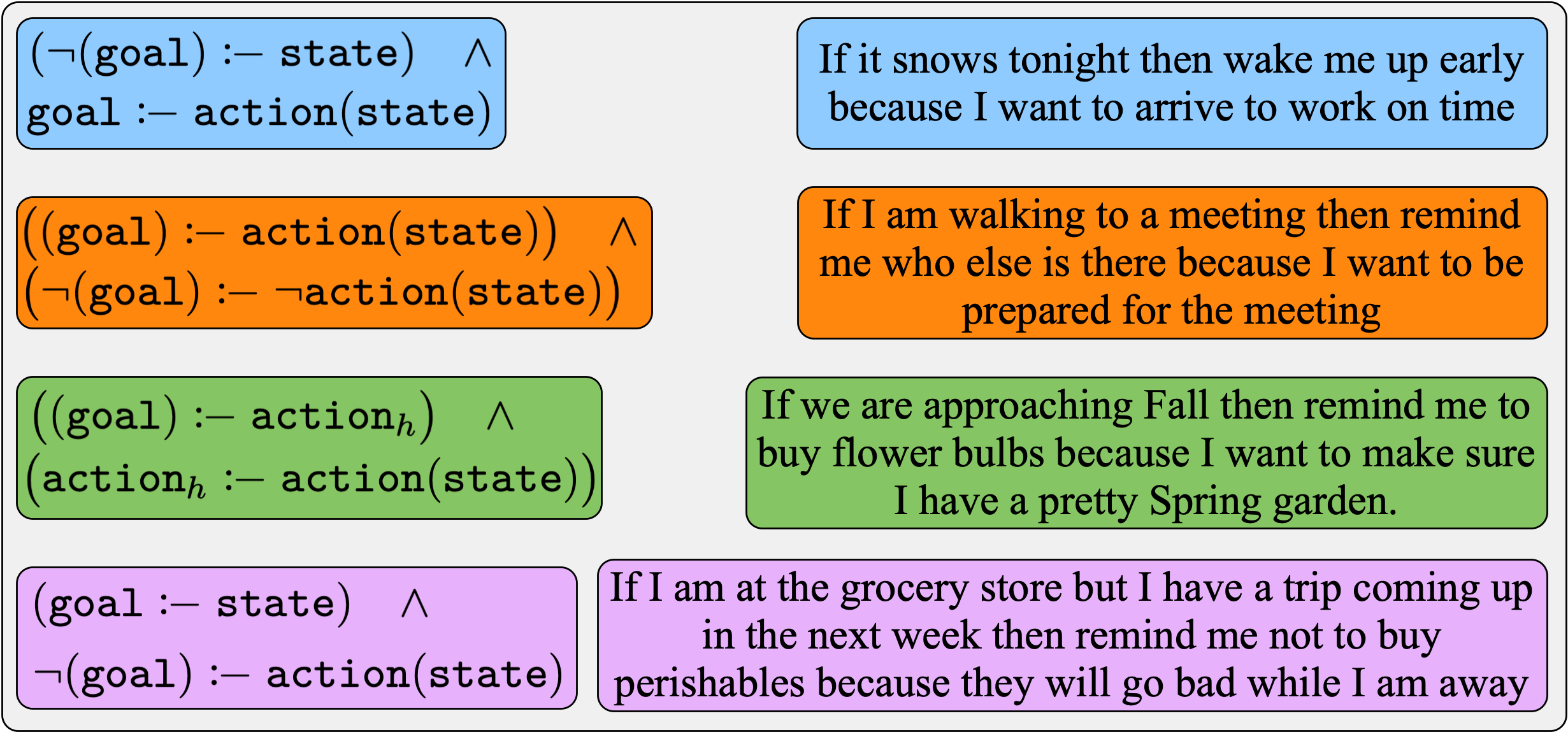}
    \caption{Logic templates and an example if-then-because command for each one. $\wedge$, $\neg$, and $\coloneq$ denote logical AND, negation, and implication, respectively. \textAction$_h$ indicates a hidden action. For example, the hidden action for the third command (green template) is ``planting flower bulbs''.}
    \label{fig:logic_template}
\end{figure}

\paragraph{Logic Templates:}
The data is partitioned into 4 color-coded reasoning logic templates that indicate how the commanded \textAction~leads to the \textGoal~when the \textState~holds. To be self-contained, we have included the table of logic templates from \citeauthor{corgi} in Figure \ref{fig:logic_template}. In the interest of space, we give one example of the blue logic template, \blueTemplate{$(\neg(\text{\textGoal}) \coloneq \text{\textState}) \wedge (\text{\textGoal} \coloneq \text{\textAction}(\text{\textState}))$}, where $\neg$ indicates negation, and $\wedge$ indicates logical AND. Under this template, the \textState~implies the negation of the \textGoal~AND the \textAction~implies the \textGoal~when the \textState~holds. An example if-then-because command that satisfies this  template is, ``If it snows tonight then wake me up early because I want to get to work on time''. Here, the \textState~of snowing (a lot) at night results in the negation of the \textGoal~and the user will not be able to get to work on time. AND if the \textAction~of waking the user up earlier is performed when the state of snowing (a lot) at night holds, the user will get to work on time. The other three follow different logic templates, but are of the same nature. We refer the reader to \citeauthor{corgi} for more details. 

\subsection{CORGI}
COmmonsense ReasoninG By Instruction (CORGI; \citet{corgi}) is the SOTA commonsense reasoning engine designed for inferring commonsense presumptions of if-then-because commands. CORGI has a neuro-symbolic reasoning module (theorem prover) that generates multi-hop reasoning trees given if-then-because commands. The neuro-symbolic reasoning module is a soft logic programming system that reasons using backward chaining (a backtracking algorithm). CORGI does not use the logic templates to do reasoning. 
One of the main limitations of CORGI is that it uses a small hand-crafted KB that is not diverse enough to reason about all the if-then-because commands in the benchmark dataset. CORGI was tested only on 10 commands in the released benchmark \citep{corgi}. Moreover, CORGI's KB is programmed in a syntax similar to Prolog \cite{colmerauer1990introduction}, but large-scale commonsense Prolog-like knowledge bases are not readily available. Therefore, in this paper we propose to use a large-scale SOTA neural commonsense KB which both extends that limited source of background knowledge and also enables the use of a new \emph{type} of knowledge source.

CORGI is equipped with a conversational interaction strategy to acquire knowledge from humans when that knowledge is missing from the KB. However, CORGI uses a static question generation strategy that often confuses the end-users. First, the parser often omits verb conjugation or subjects, resulting in grammatically incorrect questions. Second, the static question generation strategy does not always generate contextual queries. Therefore, in this paper we propose a dynamic question generation strategy that generates more relevant questions resulting in a higher quality knowledge extraction from humans. We also include a more robust parser to ensure the questions are grammatically correct.

\subsection{COMET}
COMmonsensE Transformer (\comet\cometemoji; \citeauthor{Bosselut2019COMETCT}) is a generative transformer-based commonsense KB that learns to generate rich and diverse commonsense descriptions. 
\comet~constructs commonsense KBs by using existing tuples as a seed set of knowledge on which to train. In essence, a pre-trained language model learns to adapt its learned representations to knowledge generation, producing novel high-quality tuples.

Unlike other KBs, \comet~represents knowledge implicitly in its neural network parameters and expresses knowledge through generated free-form open-text descriptions. 
This makes it well-suited for the studied reasoning task, as the if-then-because commands are also expressed in free-form text descriptions. Therefore, we can directly query \comet~for commonsense pre-conditions and post-effects of the \textState, \textAction and \textGoal. For example, ``pouring coffee'' is a commonsense pre-condition for ``drinking coffee'' (Fig \ref{fig:atomic}). 



Our system uses a \comet~model trained on two knowledge graphs, ATOMIC \cite{sap2019atomic} and ConceptNet \cite{Speer2016ConceptNet5A}. Each of these knowledge graphs consists of a collection of tuples, $\{s, r, o\}$, where $s$ and $o$ are the \emph{subject} and \emph{object} phrase of the tuple, respectively and $r \in [r^{0},r^{1},\dots,r^{\ell-1}]$ is the \emph{relation} between $s$ and $o$, and $\ell$ is the number of relations in the KB. \comet~is trained to generate $o$ given $s$ and $r$. E.g., Figure \ref{fig:comet} shows examples of the generated objects given an input subject, ``I drink coffee,'' and several relations from the ATOMIC-trained (Fig.~\ref{fig:atomic}) and ConceptNet-trained (Fig.~\ref{fig:conceptnet}) \comet{} models.

\section{\coolname: Conversational Multi-Hop Reasoner}
\coolname~(ConversationaL mUlti-hop rEasoner) is a commonsense reasoning engine that inputs if-then-because commands and outputs a reasoning chain (proof) containing if-then logical statements that indicates how the \textAction achieves the user's \textGoal when the \textState holds. \coolname~is built on top of CORGI and is triggered when CORGI fails. It consists of three components: (1) the parser, (2) the multi-hop prover (reasoner), and (3) the dialog system (Figure \ref{fig:CLUE_diagram}). The Parser takes in the command and extracts the \textState, \textAction and \textGoal from it. At the second step, the prover attempts to find reasoning chains that connect the \textAction to the \textGoal when the \textState holds. The extracted reasoning chains go through the dialog system that generates and presents contextualized questions to a human user who either validates a returned reasoning chain or contributes novel knowledge to \coolname~if none of the reasoning chains are correct.

All three components interact with \coolname's two sources of background knowledge: (I) A small handcrafted knowledge base containing logic facts and rules, $\mathcal{K}$, programmed in a logic programming language and (II) A large scale neural knowledge base, \comet, containing free-form text. During the user-interaction sessions, the system's background knowledge base $\mathcal{K}$ grows in size as new knowledge is added to it. This new knowledge is either novel information added by the user during conversational interactions, or knowledge queried from \comet~and confirmed by the user. This new knowledge will be added to $\mathcal{K}$ for future similar reasoning tasks. In what follows, we explain the three components of \coolname~in detail. 

\subsection{Parser}
Our parser extracts the \textState, \textAction, and \textGoal~as independent clauses from the input if-then-because command. We use Spacy's \cite{honnibal2017spacy} NLP tools such as POS tagging and coreference resolution to make the clauses self-contained and contextual. 

\subsection{Multi-Hop Prover}
\label{sec:prover}
\begin{figure}[t]
\centering
    \begin{subfigure}{\columnwidth}
    \includegraphics[clip, trim=0cm 9.5cm 0.5cm 0cm, width=\columnwidth]{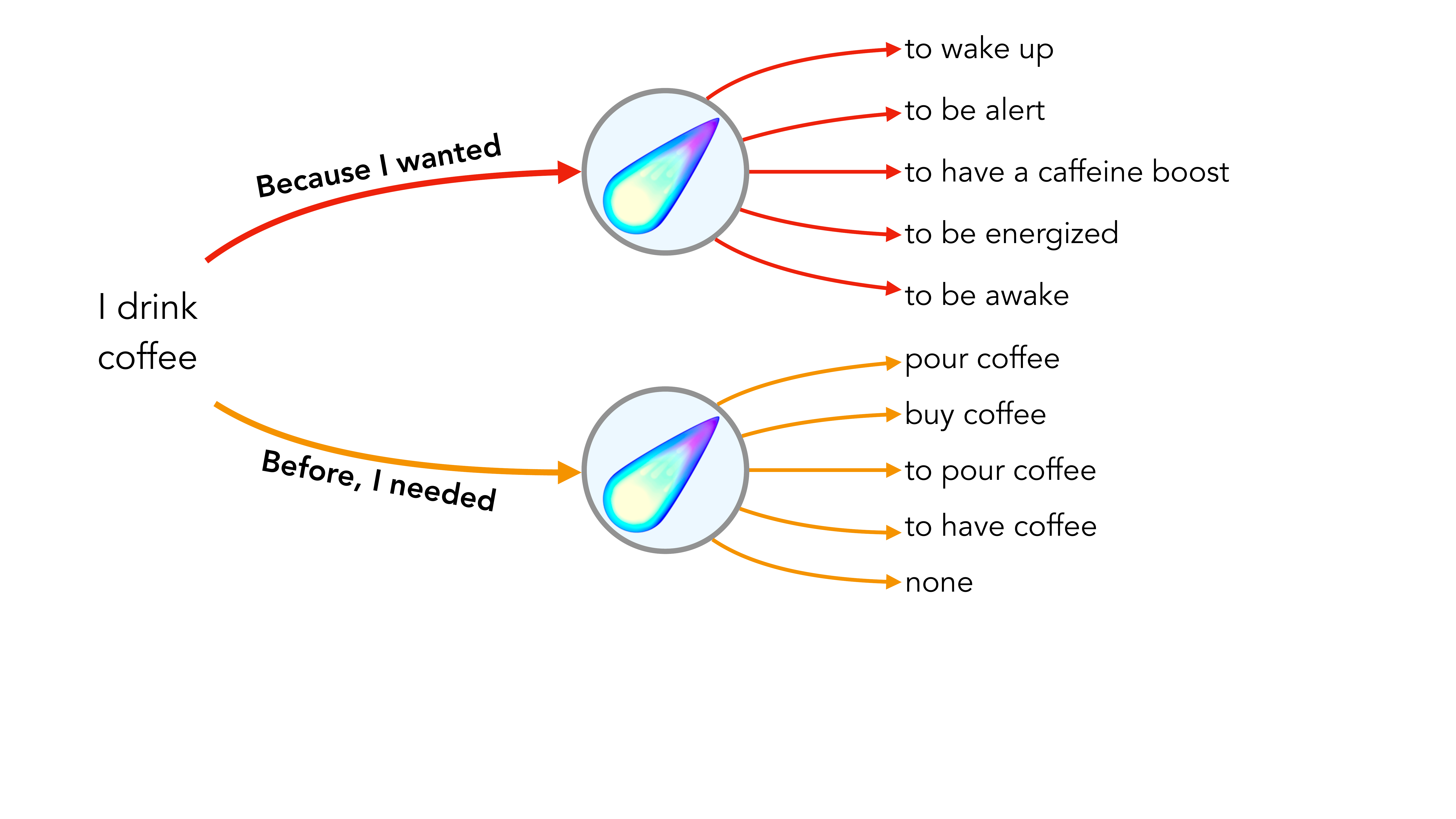}
    \caption{ATOMIC beam results for two relations \emph{Because I wanted} and \emph{Before, I needed}}
    \label{fig:atomic}
    \end{subfigure}
    \begin{subfigure}{\columnwidth}
    \includegraphics[clip, trim=0cm 9.5cm 0.5cm 0cm, width=\columnwidth]{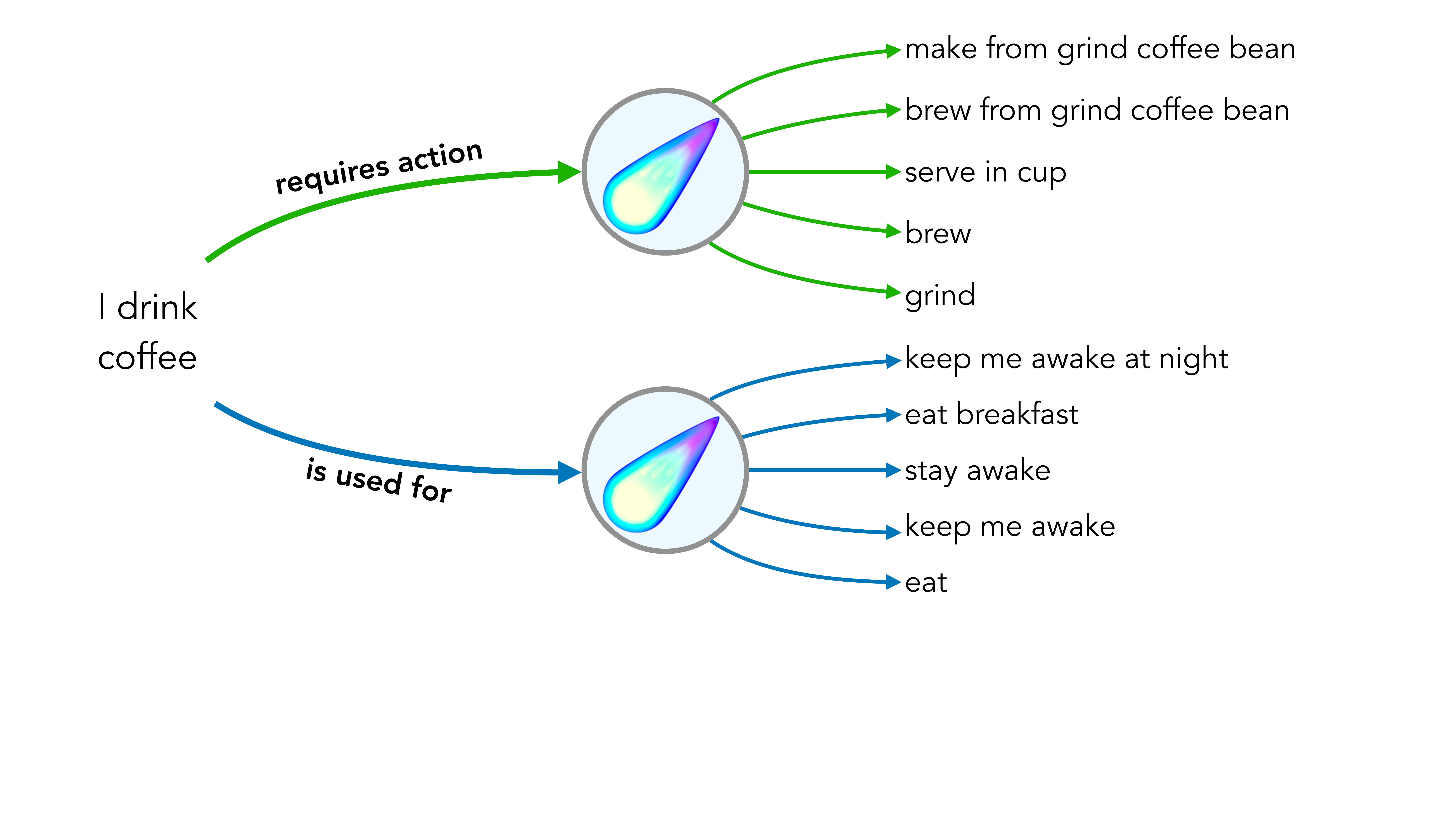}
    \caption{ConceptNet beam results for two relations \emph{requires action} and \emph{is used for}}
    \label{fig:conceptnet}
    \end{subfigure}
    \caption{\comet$_r(s)$~generations for $s=$ \emph{I drink coffee} and two different relations $r$. Our neural KB was trained on ATOMIC and ConcepNet knowledge graphs. The generations are listed on the right of the \cometemoji~block.}
    \label{fig:comet}
\end{figure}

Inspired by how Prolog generates proofs containing chains of logical rules and facts using a KB, our prover (reasoner) generates chains of natural language commonsense facts and rules to reason about an input command by iteratively querying \comet~ (details on the analogue between Prolog and our prover are in the appendix). We denote a \comet~query by \comet$(r, s)$ where $r$ is a relation and $s$ is a subject or an input natural language clause. \comet~uses beam search at decoding time and \comet$(r, s)$ outputs a list of candidate objects $O = [o^{0},o^{1},\dots,o^{b-1}]$ ordered by confidence, where $b$ indicates \comet's beam size. We categorize \comet~\emph{relation}s into two classes, namely pre-conditions and post-effects. For example, the relations ``Because I wanted'' and ``is used for'' in Fig~\ref{fig:comet} are post-effects, whereas the relations ``Before I needed'' and ``requires action'' are pre-conditions.

Let us now formally define a reasoning chain or proof extracted from our neural knowledge base. A proof consists of a chain of knowledge tuples $\{s_i, r_i, o_i\}$, where $i \in [1,N]$ and $N$ indicates the number of hops in the proof chain such that $o_{i-1}$ is \emph{semantically close} to $s_i$. (We discuss the notion of semantic closeness in the next subsection.) The search space for finding a proof chain from \comet~grows exponentially with $N$ if implemented naively. In the next subsections we explain how we prune the search space using the logic templates released with the benchmark dataset. 

The goal of our prover is to scale up CORGI's small hand-crafted knowledge base, which is programmed in a Prolog-like language. Since a large-scale commonsense KB in Prolog is not readily available, \coolname~proposes an alternative prover that enables using SOTA large-scale commonsense KBs. Moreover, \coolname's prover is consistent with CORGI's neuro-symbolic logic theorem prover (refer to the appendix for details), 
allowing us to seamlessly extend CORGI's background knowledge without requiring a large scale commonsense KB programmed in Prolog. Lastly, \coolname~ performs reasoning in a zero-shot manner since the pre-trained \comet~is not trained on the benchmark dataset used for evaluation in this paper. 


\paragraph{Semantic Closeness:} 
In order to measure semantic closeness of the \emph{object} and \emph{subject} phrases, 
we embed the sequence of tokens that make up the object of the previous tuple ($o_{i-1}$) and the sequence of tokens that make up the subject of the next tuple ($s_i$) in the proof trace. Semantic closeness, used for ranking the returned proofs, is defined as a vector cosine similarity of larger than a threshold, $\tau$.

We investigate several embedding methods to find one that best suites our multi-hop prover.
We used GloVe embeddings \cite{pennington2014glove}, BERT pre-trained embeddings \cite{devlin2019bert} and fine-tuned embeddings in \comet, which we call \emph{commonsense embeddings}. For GloVe and BERT embeddings, we compute the phrase embedding by averaging the embeddings of the tokens. For commonsense embeddings, we use the phrase embeddings returned by \comet. In the results section, we compare the outcome of these choices.

\begin{figure}[t]
    \centering
    \includegraphics[width=\columnwidth]{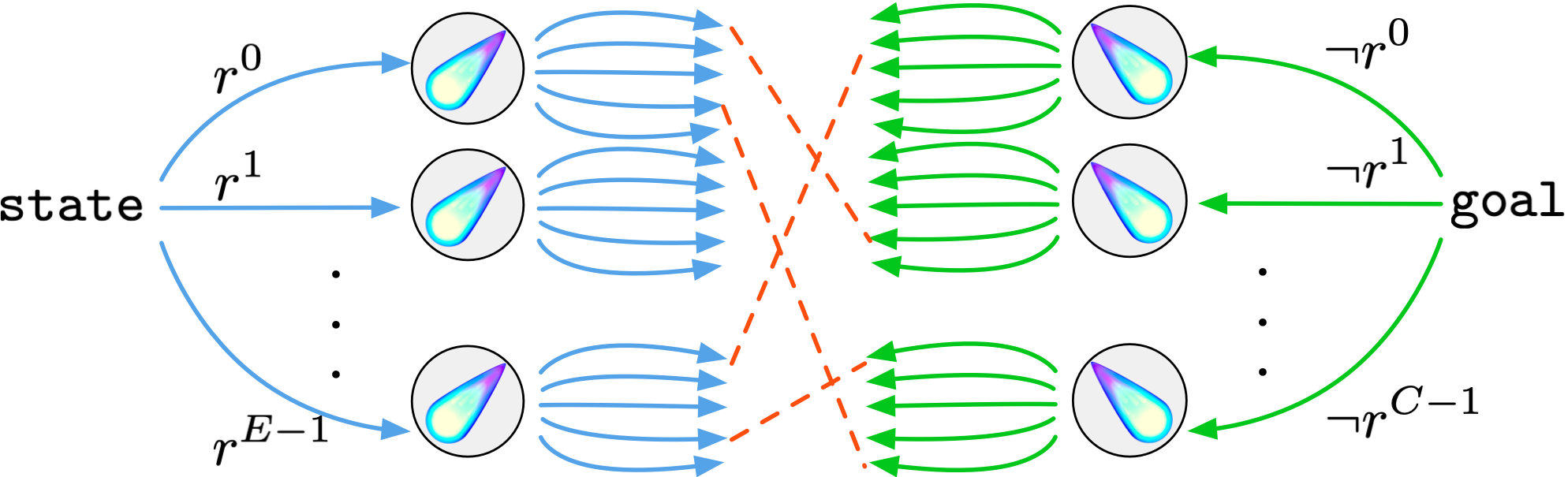}
    \caption{Bidirectional two-hop proof for the blue template \blueTemplate{$(\neg(\text{\textGoal}) \coloneq \text{\textState})$}. The dotted orange lines indicate semantic closeness. $r^{i}$'s are KB relations, $C$ is the number of pre-condition relations and $E$ is the number of post-effect relations in the KB. The $\neg$ symbol refers to relations that represent negation; for example, \emph{NotCapableOf} or \emph{NotIsA}.}
    \label{fig:blue_template}
\end{figure}

\paragraph{Pruning the Proof Search Space}
In order to form a reasoning chain (proof) for an if-then-because command using \comet, we leverage the logic templates discussed in Sec. \ref{sec:background}. All the templates consist of a conjunction of two logical implications. For each implication (\prologTerm{Head $\coloneq$ Body}), the \prologTerm{Head} and \prologTerm{Body} are given in the if-then-because command. For example, the \textGoal~and \textState~in the first implication of the purple template (\!\!\purpleTemplate{$\text{\textGoal} \coloneq \text{\textState}$}) are extracted from the input command. Therefore, in order to prove the first implication, we need to find a chain of reasoning that leads from the \textState~to the \textGoal in a series of $N$ hops. In order to do that, we either perform unidirectional or bidirectional beam search as explained below.

\paragraph{\emph{Unidirectional Beam Search:}} Here, for a given number of hops $N$, we first construct all beam results $O_1=$\comet$(r_{e}, s_1)$ where $s_1$ is the natural language clause that corresponds to the \prologTerm{Body} of the implication in the logic template, and for all $r_e$ that are post-effect \comet~relations. 
We continue to query \comet($r_e, o_n^j$)~recursively in a breadth-first manner for $\forall~o_n^j \in O_n$ where  $n \in [1,\dots,N]$ is the hop index. In each hop, we only continue the query for the top $K$ results ranked by the semantic closeness of the returned beam result $o_n^j$ and the implication's \prologTerm{Head}, where $K$ is the search's beam size. At the $N^{\text{th}}$ hop, the prover returns proof chains for which $o_N$ is semantically close to the implication's \prologTerm{Head}, which corresponds to the \textGoal.

\paragraph{\emph{Bidirectional Beam Search:}} Here, we construct all possible beam results of \comet$(r_e, s_1)$, where $s_1$ is the natural language clause that corresponds to the implication's \prologTerm{Body}, for all $r_e$ that are post-effect relations as well as \comet$(r_c, o_N)$, where $o_N$ is the natural language clause corresponding to the \prologTerm{Head} of the implication in the logic template, for all $r_c$ that are pre-condition relations. 
The proof succeeds when two intermediate beam results are semantically close in the beam search path from either direction (Fig \ref{fig:blue_template}).

\subsection{Dialog System}
The reasoning chains obtained by the prover are passed to the Dialog System, which has two main goals. The first is to confirm the prover obtained valid reasoning chains by asking the user if they think the automatically recovered proofs are ``correct'' from a commonsense stand point (Fig \ref{fig:dialog}). The necessity of this was also confirmed in \citet{Bosselut2019COMETCT}'s human evaluation studies. The second is to overcome the problem of missing knowledge when the user rejects all the proofs returned by the prover. We introduce the dialog generators responsible for fulfilling each of the above 
goals in what follows.


\paragraph{Humans as Knowledge Evaluators}
The dialog generator 
confirms the returned \comet~proofs with humans before adding it as background knowledge to $\mathcal{K}$. In order to do this, it chooses the top 5 proofs with the highest similarity scores and presents them as candidates to the human user to choose from. Our study shows that these multiple choices not only help confirm \comet~results but they also provide guidance to users as to what information the system is looking for. 

As shown in Figure \ref{fig:dialog}, five explanations for the question are returned by \coolname. The user chooses one and provides an explanation as to why he/she chose that option. The explanation is in open-domain text, formatted as a series of if-then statements. This is because if-then formatted explanations can be easily 
parsed using our parser. 
In this step, the question asked from users is contingent on which logic template the if-then-because command follows. The dialog system's flowchart as well as an example dialog are in the appendix.

\paragraph{Humans as Novel Knowledge Contributors}
We also use human interaction to acquire novel knowledge that does not exist in the background knowledge bases, $\mathcal{K}$ and \comet. 

When faced with missing knowledge in $\mathcal{K}$, \coolname~uses the same technique from \citet{corgi} with a rephrased, more comprehensive question, ``what ensures $\langle$ \textGoal$\rangle$?'' resulting in higher quality feedback. 
If the user's response to the multiple-choice question is ``None of the above'', it indicates that \comet~has missing knowledge. \coolname~then asks the user for an explanation and adds the new knowledge to $\mathcal{K}$ and runs CORGI's reasoning module to construct a proof. 
It is worth noting that since $\mathcal{K}$ is orders of magnitudes smaller than SOTA knowledge bases, growing it with novel knowledge does not introduce any scalability issues.
\section{Experiments and Results}
Here we discuss our experimental setup, evaluation method, baselines and results. 
We work with 132 out of the 160 commands from the benchmark dataset \shortcite{corgi} that fall under three logic templates blue, orange and green (Figure \ref{fig:logic_template}). The red template contains commands for which there is no unifying logic template. Therefore, we cannot use it.

\paragraph{Evaluation:}
We do not use automated evaluation metrics and instead use human evaluations for two reasons. First, 
there are currently no metrics in the literature that assess whether the returned reasoning chains are ``correct'' from a commonsense perspective. 
Second, evaluating dialog systems is challenging. It is debated that metrics such as BLEU \cite{papineni2002bleu} and perplexity often fail to measure true response quality \cite{liu2016not,li2016deep}. 

\paragraph{Experiments:}
In the first experiment, we evaluate our multi-hop prover in isolation and without conversational interactions. The human evaluators in this study are expert evaluators. In our second experiment, we test \coolname's performance end-to-end with non-expert human users and investigate the efficacy of the conversational interactions.
In order to be comparable with \citet{corgi}'s study, we use the knowledge base of commonsense facts and rules, $\mathcal{K}$, released with the dataset. It contains 228 facts and rules. 
Our \comet~model is pre-trained on knowledge graphs ConceptNet 
and ATOMIC. 
We used \comet's open-source code\footnote{https://github.com/atcbosselut/comet-commonsense} for training with the hyper-parameters reported there \citep{Bosselut2019COMETCT}.


\subsection{Multi-hop Prover Evaluation}
As shown in Figure \ref{fig:logic_template}, each logic template consists of a conjunction of two logical implication statements. We use the terminology \emph{end-to-end} proof to refer to proving both of the implications in the template and \emph{half-proof} to refer to proving one. 

Table \ref{tab:n_hop} 
presents the number of proved logical implications \textGoal $\coloneq$ \textAction, with respect to the number of hops 
using unidirectional and bidirectional beam search. Expert-verified proofs refer to reasoning chains with similarity score of at least $0.8$ that are validated by a human evaluator. 
The similarity threshold of $0.8$ was tuned offline on a small subset of the benchmark dataset 
and picked from the following list $[0.7, 0.8, 0.85, 0.9]$. Automated proofs are the portion of the human evaluated proofs for which the highest scoring proof is the verified one. As shown, the number of automated proofs almost doubles when an expert human evaluator validates the proofs. 
This indicates that the model benefits from \emph{human knowledge evaluators}. An instance of this scenario is shown in Figure \ref{fig:dialog} where the human user chooses the second ranking candidate as the correct proof. As expected, a portion of the commands cannot be proved using \comet~alone even with human evaluators. This indicates that we are encountering missing knowledge. Therefore, there is need for \emph{humans as novel knowledge contributors}. 
Moreover, Table \ref{tab:n_hop} shows that bidirectional beam search is more successful than unidirectional search for lower hops greater than $1$. This is because there is a higher chance of finding a good scoring match when the two directions meet due to an extended search space. For the same reason, the number of successful proofs drop when the number of hops is increased beyond a certain point. Please note, in bidirectional beam search $N=1$ is not applicable because the smallest number of hops extracted for bidirectional beam search is 2. Moreover, if a statement is proved with lower number of hops, we do not prove it with a higher $N$. Therefore, the number of hops needed for proving a certain statement is not predefined, and is rather chosen based on the best semantic closeness among all extracted hops at test time. 
The automated half-proofs obtained from the prover are listed in Tables \ref{tab:half_proof_ex_p1} and \ref{tab:half_proof_ex_p2} in the Appendix.


\begin{table}[t]
\centering
\caption{Number of hops ($N$) required to obtain a half-proof for the implication \textGoal $\coloneq$ \textAction for two proof search strategies and 132 commands. Similarity score is computed using GloVe embeddings.}
\label{tab:n_hop}
\resizebox{\columnwidth}{!}{%
\begin{tabular}{lcccccc}
\toprule
\backslashbox{Pruning}{N} & & 1 & 2 & 3 & 4 & 5
\\ \midrule
\multirow{2}{*}{\thead{Unidirectional\\beam search}} & \# Expert-verified proofs & 50 & 21 & 13 & 2 & 1 \\
 & \# Automated proofs & 26 & 10 & 6 & 1 & 1 \\
\midrule
\multirow{2}{*}{\thead{Bidirectional\\beam search}} & \# Expert-verified proofs & N/A & {\bf 67} & 16 & 9 & 3 \\
 & \# Automated proofs & N/A & {\bf 42} & 6 & 0 & 0 \\
\bottomrule
\end{tabular}
}
\end{table}
\begin{table}[t]
\centering
\small
\caption{Number of successful unidirectional half-proofs for the implication \textGoal $\coloneq$ \textAction for a given number of hops ($N$) among the top $k$ proof candidates for 132 commands. Similarity score is computed using GloVe embeddings.}
\label{tab:rank_k}
\resizebox{0.6\columnwidth}{!}{%
\begin{tabular}{lcccccc}
\toprule
\backslashbox{k}{N} & 1 & 2 & 3 & 4 & 5
\\ \midrule
1 & 26 & 10 & 6 & 1 & 1 \\
2 & 32 & 12 & 7 & 1 & 1 \\
3 & 35 & 12 & 9 & 1 & 1 \\
4 & 37 & 14 & 9 & 1 & 1 \\
5 & 39 & 14 & 10 & 1 & 1 \\
\bottomrule
\end{tabular}
}
\end{table}

In Table \ref{tab:rank_k}, we present the number of expert-verified half-proofs achieved with different number of hops ($N$) within the top $k$ ranked results. Note that we exclude successful half-proofs for higher degrees of $N$ if there exists a proof with fewer hops, and we include a proof as long as the verified result is within the top $k$ results. Table \ref{tab:rank_k} shows that the majority of successful half-proofs are achievable in $N\leq3$.

In Table \ref{tab:comet_auto_full_proofs}, we present the number of obtained full proofs for 2-hop bidirectional beam search (the best result from Table \ref{tab:n_hop}) using GloVe embeddings broken down by logic templates. Although \comet~is successful at finding half proofs, the success rate decreases when two implications need to be proven in conjunction. As shown, \comet~can (fully) prove $12.8\%$ of the if-then-because commands. This emphasizes the necessity of using human conversational interactions for obtaining full proofs. Therefore, if \comet~succeeds at proving half of the template, there is a chance to prove the other half with the help of a human user. This is because most of the commands belong to the long tail of tasks for which there is too little knowledge available in any commonsense KB.


Table \ref{tab:half_proofs} compares the number of successful proofs obtained using GloVe, BERT and commonsense embeddings. 
As shown, GloVe embeddings perform better than the others since commonsense and BERT embeddings perform poorly when there is little surrounding context for the words. 
The returned \emph{objects} by \comet~tend to have a few number of tokens, (around 1-2). Therefore, the similarity scores assigned by BERT and commonsense embeddings either result in more false positives or in pruning out good candidates. For example, BERT and commonsense embeddings both tend to assign a lower similarity score to ``I sleep'' and ``sleep'' than single token words like ``ski'' and ``spoil''. But this is not an issue with GloVe embeddings. Therefore, in all the other experiments we have used GloVe embeddings. The effect of the embeddings is amplified as the number of hops increases. For example in Table \ref{tab:n_hop}, The number of expert-verified 2-hops proofs drops from 21 to 5 if we use BERT embeddings.

\begin{table}[t]
\centering
\caption{Number of End-to-End Proofs per Logic Template for 2-hop bidirectional beam search.}
\label{tab:comet_auto_full_proofs}
\resizebox{\columnwidth}{!}{%
\begin{tabular}{llccc}
\toprule
& Logic Template & Successful & False Positive & Total Count \\ \midrule
\multirow{ 3}{*}{\thead{Automated \\proof}} & orange & 1 & 0 & 50 \\
&blue & 6 & 1 & 65 \\
&green & 6 & 2 & 17 \\
\midrule
\multirow{ 3}{*}{\thead{Expert\\evaluated \\proof}} & orange & 1 & N/A & 50 \\
&blue & 7 & N/A & 65 \\
&green & 9 & N/A & 17 \\
\bottomrule
\end{tabular}
}
\end{table}

\begin{table}[t]
\centering
\caption{Number of half-proofs using bidirectional beam search: evaluated by expert human evaluators.}
\label{tab:half_proofs}
\resizebox{0.9\columnwidth}{!}{%
\begin{tabular}{lcc}
\toprule
Embedding Space & \textGoal $\coloneq$ \textAction & $\neg$ \textGoal $\coloneq$ \textState \\ 
\midrule
GloVe & 67 & 17 \\
BERT & 52 & 14 \\
Commonsense &59 & 12 \\
\bottomrule
\end{tabular}
}
\end{table}

\subsection{User Study}
To assess \coolname's end-to-end performance, we ran a user study in which human users engaged in a conversational interaction with \coolname~and answered its prompts. We collected a total of 700 dialogues from this study and report the percentage of the if-then-because commands that were successfully proved (end-to-end) as a result of the conversational interactions in Table \ref{tab:userstudybreakdown}. Our users worked on a total of 288 if-then-because commands that fit the blue, orange, and green logic templates. We had 129 unique commands and 29 participants in the study with at least 2 users per command. The 129 commands used in this study are a subset of the 132 commands (half-proofs) in Table \ref{tab:n_hop}. The remaining 3 statements were excluded because either the statements were longer than COMET's maximum input token size (for the full-proof) or they did not trigger a CORGI/CLUE interaction so it does not make sense to include them in the user-study. Each participant worked on 9-10 statements taking approximately 30-40 minutes to complete them all. The participants were undergraduate, graduate or recently graduated students with a background in computer science or engineering. Please note that no skills beyond reading is required for interacting with \coolname; no particular domain expertise is needed either because most humans possess common sense regarding day-to-day activities.


The users contributed a total number of 70 novel knowledge tuples to our background knowledge base and validated 64.86\% of the proofs extracted using our multi-hop prover. 
\textbf{}\coolname~is built on top of CORGI,\footnote{https://github.com/ForoughA/CORGI} and is triggered only if CORGI fails. Therefore, the 19 and 4 commands proved by \coolname~are not provable by CORGI; the commands proved by CORGI in each row is indicated in the parentheses. Therefore, \coolname~successfully increases the reasoning success rate by 35\%. Please note that the absolute gain in terms of success percentage compared with CORGI is 8\%.

We did not experience any inconsistencies between user responses in the study because users received the same goal for a given command. It is difficult to generate contradictory proofs when the context is the same. However, proofs among different users do not have to be identical. As long as the proof fulfills the criteria and is ``correct'' from a commonsense perspective, it holds. Every user has different preferences, linguistic tendencies, and living habits, which can affect the generated proof. Consequently, proofs are tailored to users. Also, showing the five multiple choice options helps users understand what kind of knowledge \coolname~is looking for, improving their responses.




\begin{table}[t]
\centering
\caption{User-study results. Commands not provable by CORGI trigger a \coolname~dialog, so commands proved by \coolname~are in addition to CORGI's. The numbers in the parentheses in proved column is the number of statements that CORGI proved.}
\label{tab:userstudybreakdown}
\resizebox{0.9\columnwidth}{!}{%
\begin{tabular}{cccc}
\toprule
Dialog Phase & \thead{Proved} & \thead{Tried} & \thead{Proved/Tried} \\
\midrule
CORGI & 65 & 288 & 0.2256 \\
\coolname[orange]
 & 17(+27) & 126 & 0.1349 \\
 \coolname[green] & 2(+0) & 17 & 0.1176 \\
\coolname [blue]
& 4(+38) & 145 & 0.0276 \\
\bottomrule
\end{tabular}
}
\end{table}

\section{Related Work}
Efforts in developing commonsense reasoning started as early as the foundation of the field of artificial intelligence \cite{grice1975logic,winograd1972understanding,davis2015commonsense, minskyframework}. 
Commonsense reasoning is becoming more prominent as computers increase their interactions with us in our daily lives. For example, conversational agents such as Alexa, Siri, Google Home and others have very recently entered our daily lives. However, they cannot currently engage in natural sounding conversations with their human users mainly due to lack of commonsense reasoning. Moreover, they operate mostly on a pre-programmed set of tasks. On the other hand, instructable agents \cite{azaria2016instructable,labutov2018lia,li2018appinite,li2017programming,li2017sugilite,guo2018dialog,mohan2014learning,mininger2018interactively,mohan2012acquiring}, can be taught new tasks through natural language instructions/demonstrations. One of the challenges these bots face is correctly grounding their natural language instructions into executable commands. 

Our approach addresses a new reasoning task proposed by \citet{corgi} that contains commands given to an instructable agent satisfying a general template. 
In contrast to this challenging task and TimeTravel \cite{qin2019counterfactual}, most commonsense reasoning benchmarks have traditionally been designed in a multiple choice manner. Moreover, they are not typically targeted at conversational agents. 
Refer to \citet{storks2019commonsense} and \citet{corgi} for a comprehensive list of commonsense reasoning benchmarks.

Our commonsense reasoning engine uses a SOTA neural knowledge model, \comet~\cite{Bosselut2019COMETCT}, as an underlying source of commonsense knowledge. \comet~is a framework for constructing knowledge bases from transformer-based language models. In contrast to previous automatic knowledge base construction methods that rely on semi-structured \cite{suchanek2007yago,hoffart2013yago2,auer2007dbpedia} and unstructured \cite{dong2014knowledge,carlson2010toward,mitchell2018never,nakashole2012patty} text extraction, \comet~uses transfer learning to adapt language models to generate knowledge graph tuples by learning on examples of structured knowledge from a seed KB. 
\comet~was recently used for persona-grounded dialog for chatbots \cite{majumder2020like}. 

\section{Conclusions}
We introduce the ConversationaL mUlti-hop rEasoner (\coolname) for commonsense reasoning in conversational agents. \coolname~uses a neural commonsense KB and symbolic logic rules to perform multi-hop reasoning. It takes {\bf if-(state), then-(action), because(goal)} commands as input and returns a multi-hop chain of commonsense knowledge, indicating how the \textAction  leads to the \textGoal  when the \textState  holds. The symbolic logic rules help significantly reduce the multi-hop reasoning search space and improve the quality of the generated commonsense reasoning chains. We evaluate \coolname~with a user study with human users.

\section*{Acknowledgments}
Tom Mitchell is supported in part by AFOSR under grant FA95501710218. Antoine Bosselut and Yejin Choi gratefully acknowledge the support of DARPA under No. N660011924033 (MCS), JD.com, and the Allen Institute for AI.

\clearpage

\section*{Appendix}

\section*{Multi-hop Prover Details}
Prolog is a logic programming language that consists of a set of predicates. A predicate has a name (functor) and a set of $N > 0$ arguments. For example, \prologTerm{get(i, work, on\_time)} is a predicate with functor \prologTerm{get} and 3 arguments. Predicates are defined by a set of logical rules or Horn clauses (\prologTerm{$\text{Head} \coloneq \text{Body}$}) and facts (\prologTerm{$\text{Head}$}), where \prologTerm{Head} is a predicate, \prologTerm{Body} is a conjunction of predicates, and \prologTerm{$\coloneq$} is logical implication. Prolog uses backward-chaining to logically reason about (prove) an input query, represented by a predicate. From a high level, a proof consists of a chain of logical facts and rules available in the background KB. Inspired by this prover, we extend this to free-form text proofs in this paper.

This chain-structured definition of a proof is inspired by a Prolog proof. A knowledge tuple here is analogous to a logical rule in Prolog (\prologTerm{Head $\coloneq$ Body}) where the \emph{subject} and \emph{object} correspond to either the \prologTerm{Body}, or the \prologTerm{Head} depending on the  \emph{relation}. We categorize \comet~\emph{relation}s into two classes, namely pre-conditions and post-effects. If the \emph{relation} is a post-effect, the \emph{object} is the \prologTerm{Head} and the \emph{subject} is the \prologTerm{Body}. If the \emph{relation} is a pre-condition, the reverse is true. For example, the relations ``Because I wanted'' and ``is used for'' in Fig~\ref{fig:comet} are post-effects, whereas the relations ``Before I needed'' and ``requires action'' are pre-conditions.
The \emph{subject} (\prologTerm{Body}) consists of a single predicate (instead of a conjunction of predicates in Prolog). Since there is no conjunction in the \prologTerm{Body} of the rules, the logical proof reduces to a chain (as opposed to a proof tree in Prolog). Moreover, the semantic closeness of $o_{i-1}$ and $s_i$ is inspired by the unification operation \cite{colmerauer1990introduction} in Prolog and is analogous to the soft unification operation of CORGI's neuro-symbolic theorem prover. 
It is worth noting that this analogue falls short of Prolog's variable grounding and is an interesting avenue for our future work.

\section*{Dialogue System Details}
\begin{figure}[h]
    \centering
    \includegraphics[width=\columnwidth]{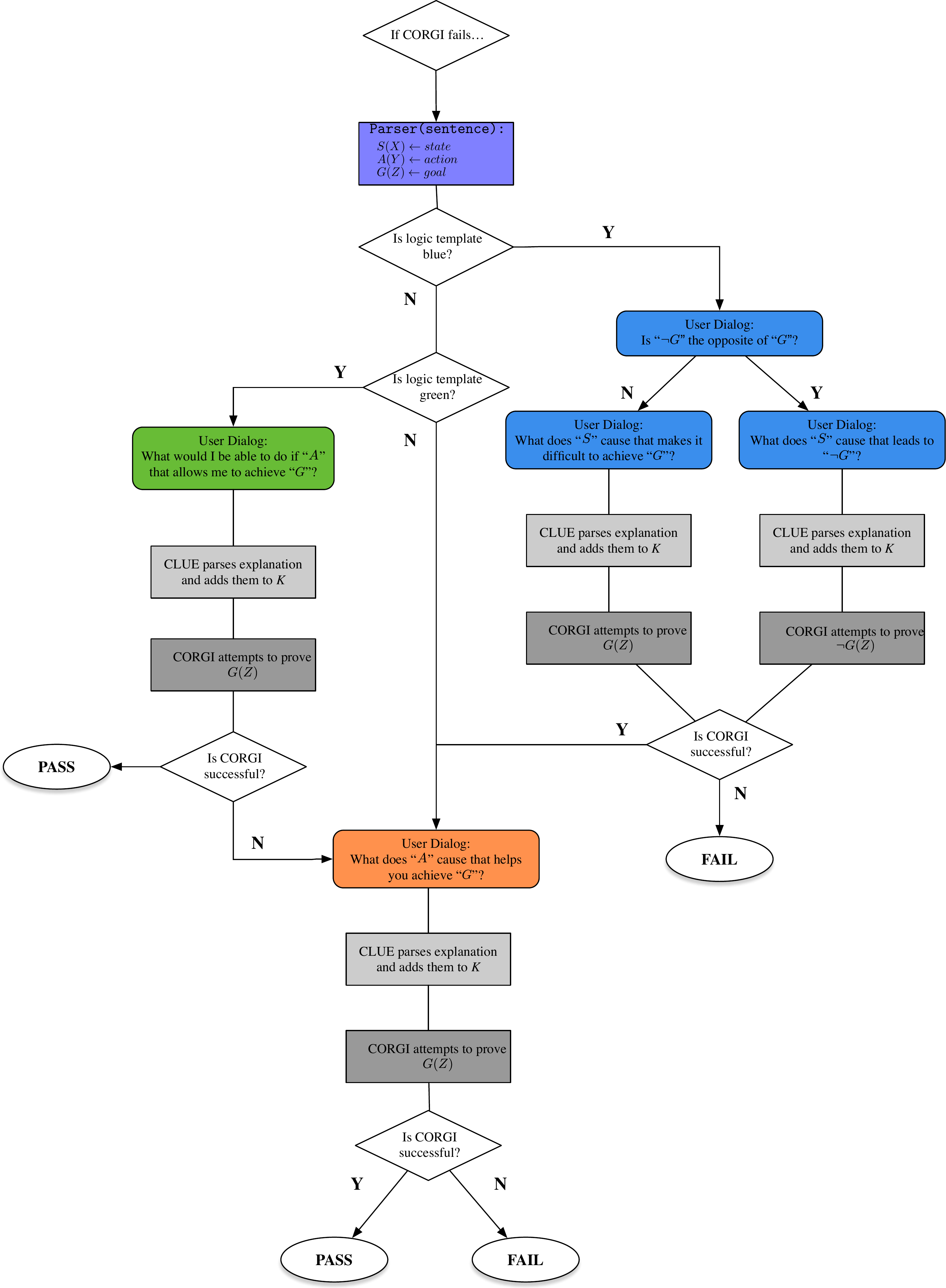}
    \caption{Full Control Flow Diagram for the Dialogue System in Figure \ref{fig:CLUE_diagram}. 
    }
    \label{fig:control_flow}
\end{figure}

The control flow of \coolname's dialog system in Figure \ref{fig:CLUE_diagram} is shown in detail in Figure \ref{fig:control_flow}. The dialog system uses the logic templates and the results returned by \comet{} to interact with the human user and ask questions.

For example, consider proving the first implication of the blue template (\!\!\!\!\!\blueTemplate{$\neg$ \textGoal $\coloneq$ \textState}) for the command ``If it snows tonight then wake me up early because I want to get to work on time''. \coolname~first tries to find the negation of the \textGoal (get to work on time) by querying \comet$(r, \text{\textGoal})$ for relations ($r$) indicating negation (such as \emph{NotCapableOf} and \emph{NotIsA}). \coolname~then picks the highest ranking returned \comet~statement (I am late) and asks the user ``is $\langle \neg$ \textGoal $\rangle$ the opposite of $\langle$ \textGoal $\rangle$?(y/n)'' (Is ``I am late'' the opposite of ``I get to work on time''?(y/n)). If the user responds `yes', then \coolname~asks ``what does $\langle$ \textState $\rangle$ cause that leads to $\langle \neg$ \textGoal $\rangle$?'' (what does ``it snows tonight'' cause that leads to ``I am late''?) and expects an explanation from the user in response. If the user responds `no' to the first question, then \coolname~asks ``what does $\langle$ \textState $\rangle$ cause that makes it difficult to achieve $\langle$ \textGoal $\rangle$?'' (What does ``it snows tonight'' cause that makes it difficult to achieve ``I get to work on time''?) and expects an explanation from the user in response. The reason for querying \comet~for the negated \textGoal~is that negation in Prolog is implemented based on negation as failure. Therefore, to be consistent, \coolname~converts the \textGoal to its negated statement and proves that instead.

\section*{Extended Experiments}

\begin{table*}[t]
\caption{Half Proofs \textGoal $\coloneq$ \textAction, Part 1. The half-proofs are obtained using our proposed prover and are obtained using bidirectional pruning. The input if-then-because commands are listed in the first column. The semantic clossenesse scores are obtained with GloVe embeddings. $\mathbf{(r_1, r_2)}$ on the third column are relations tuples on the final proof path obtained from the \comet$\mathbf{(r_1,\text{\textAction})}$ and \comet$\mathbf{(r_2,\text{\textGoal})}$ queries.}
\label{tab:half_proof_ex_p1}
\resizebox{\textwidth}{!}{ 
\begin{tabular}{|l|l|l|l|}
\hline
\multicolumn{1}{|c|}{\textbf{Commands}} & \multicolumn{1}{c|}{\thead[c]{\textbf{Semantic}\\ \textbf{Closeness Score}}} & \multicolumn{1}{c|}{\thead[c]{\textbf{$\mathbf{(r_1,r_2)}$}}} 
& \multicolumn{1}{c|}{\textbf{\Big(\comet$\mathbf{(r_1,\text{\textAction})}$, \comet$\mathbf{(r_2,\text{\textGoal})}$\Big)}} \\ \hline
\thead[l]{if i have an upcoming exam \\ then remind me to prepare 3 days ahead \\ because i want to prepare for it} & 1.0 & (CausesDesire, HasSubevent) & (prepare for exam, prepare for exam) \\ \hline
\thead[l]{if the air temperature is forecast to be warmer than 70 tonight \\ then remind me to turn on the air conditioner \\ because i want to stay cool} & 1.0 & (CapableOf, CreatedBy) & (cool air, cool air) \\ \hline
\thead[l]{if i haven't been to the gym for more than 3 days \\ then remind me to go to the gym \\ because i want to stay fit} & 0.9220791 & (CapableOf, HasPrerequisite) & (work out, work out regularly) \\ \hline
\thead[l]{if i am going to school \\ then remind me to take my office keys with me \\ because i want to be able to unlock my office door} & 1.0 & (CausesDesire, CausesDesire) & (go to work, go to work) \\ \hline
\thead[l]{if the temperature is going to be below 40 degrees in the evening but above 40 degrees in the morning \\ then remind me to bring a jacket \\ because i want to stay warm on my commute} & 1.0 & (CausesDesire, Prerequisite) & (wear jacket, wear jacket) \\ \hline
\thead[l]{if i get an email from my boss about our upcoming \\ deadline then notify me about the email because i want to read the email} & 1.0 & (Causes, CausesDesire) & (open email, open email) \\ \hline
\thead[l]{if my calendar is clear today, then remind me to go to \\gym in the afternoon, because i want to keep myself healthy} & 1.0 & (Desires, Desires) & (exercise, exercise) \\ \hline
\thead[l]{if i start using google maps to go home \\ then tell alexa to turn on the heat \\because i want my home to be warm when i arrive} & 1.0 & (Desires, CreatedBy) & (heat, heat) \\ \hline
\thead[l]{if there is heavy traffic in the route that i use to office \\then remind me to leave early \\because i want to reach office on time} & 0.9641539 & (CausesDesire, CausesDesire) & (go to work early, go to work) \\ \hline
\thead[l]{if papers related to what i'm working on are posted on the proceedings of any nlp or ml conference \\ then tell me about the papers immediately \\ because i want to stay up-to-date on current research} & 1.0 & (HasSubevent, CausesDesire) & (read, read) \\ \hline
\thead[l]{if i'm not in bed at 12pm \\then remind me to go to bed \\because i want to go to bed early} & 1.0 & (Desires, CreatedBy) & (sleep, sleep) \\ \hline
\thead[l]{if a new paper related to what i'm working on is posted on arxiv \\ then notify me about the new paper immediately \\ because i want to stay up-to-date on current research} & 1.0 & (HasSubevent, CausesDesire) & (read, read) \\ \hline
\thead[l]{if the price of something i want to buy drops \\then notify me about the price drop \\because i want to buy it when the price is low} & 0.9652006 & (MotivatedByGoal, CausesDesire) & (buy something, buy something else) \\ \hline
\thead[l]{if there is heavy traffic on my current commute path \\ then give me a less congested path \\ because i want to minimize my driving time} & 1.0 & (MotivatedByGoal, HasSubevent) & (i drive fast, i drive fast) \\ \hline
\thead[l]{if i have more than ten unread emails \\then set an one hour email replying event on calendar \\because i want to be responsive to emails} & 1.0 & (CausesDesire, HasPrerequisite) & (send email, send email) \\ \hline
\thead[l]{if i have set an alarm for taking my pills \\ then make sure the alarms are off after i have finished my pills \\ because i want to make sure i don't have false alarms} & 1.0 & (HasPrerequisite, HasPrerequisite) & (turn off alarm clock, turn off alarm clock) \\ \hline
\thead[l]{if the weather temperature forecast in the next 10 days is above 30 degrees celsius \\ then remind me to turn the heater off \\ because i don't want to make the house warm} & 0.93141675 & (MotivatedByGoal, HasSubevent) & (it be cold, it get cold) \\ \hline
\thead[l]{if i am driving home and i have an email about a grocery shopping list \\ then remind me to stop at the grocery store \\ because i want to buy the items on my grocery shopping list} & 0.99999994 & (CapableOf, UsedFor) & (buy grocery, buy grocery) \\ \hline
\thead[l]{if there is a sale on sketchbooks between now and august \\then notify me about the sale \\because i need to get sketchbooks for my fall class} & 0.8800874 & (CausesDesire, HasPrerequisite) & (buy something, buy them) \\ \hline
\thead[l]{if i have set an alarm for a time between 2am-8am on weekends \\then notify me that i have an alarm set \\because i want to correct the alarm} & 0.93820596 & (CreatedBy, CreatedBy) & (turn off alarm, turn off alarm clock) \\ \hline
\thead[l]{If the air temperature is forecast to be colder than 40 degrees \\then tell me to close the windows \\because I want to stay warm} & 1.0 & (Desires,  HasPrerequisite) & (get warm, get warm) \\ \hline
\thead[l]{if the air temperature is forecast to be warmer than 70 tonight \\then remind me to turn on the air conditioner \\because i want to stay cool} & 0.9137391 & (UsedFor, UsedFor) & (cool down room, cool down) \\ \hline
\thead[l]{if there is a natural disaster back at home \\then remind me to donate money \\because i want to give back to my community} & 0.9001999 & (CapableOf, Causes) & (i donate money, i give money) \\ \hline
\end{tabular}
}
\end{table*}

\begin{table*}[h]
\caption{Half Proofs \textGoal $\coloneq$ \textAction, Part 2. The half-proofs are obtained using our proposed prover and are obtained using bidirectional pruning. The input if-then-because commands are listed in the first column. The semantic clossenesse scores are obtained with GloVe embeddings. $\mathbf{(r_1, r_2)}$ on the third column are relations tuples on the final proof path obtained from the \comet$\mathbf{(r_1,\text{\textAction})}$ and \comet$\mathbf{(r_2,\text{\textGoal})}$ queries.}
\label{tab:half_proof_ex_p2}
\resizebox{\textwidth}{!}{ 
\begin{tabular}{|l|l|l|l|}
\hline
\multicolumn{1}{|c|}{\textbf{Commands}} & \multicolumn{1}{c|}{\thead[c]{\textbf{Semantic}\\ \textbf{Closeness Score}}} & \multicolumn{1}{c|}{\thead[c]{\textbf{$\mathbf{(r_1,r_2)}$}}} 
& \multicolumn{1}{c|}{\textbf{\Big(\comet$\mathbf{(r_1,\text{\textAction})}$, \comet$\mathbf{(r_2,\text{\textGoal})}$\Big)}} \\ \hline
\thead[l]{if an author i like is doing a reading in my city \\then let me know about the reading \\because i want to see them} & 0.8933883 & (HasPrerequisite, HasPrerequisite) & (go to bookstore, go to store) \\ \hline
\thead[l]{if i have an upcoming bill payment \\then remind me to pay it \\because i want to make sure i avoid paying a late fee} & 1.0 & (CausesDesire, Desires) & (pay bill, pay bill) \\ \hline
\thead[l]{If it snows tonight \\then wake me up early \\because I want to get to work early} & 0.95353657 & (Causes, Causes) & (i go to work early, get to work early) \\ \hline
\thead[l]{if i have a meeting \\then remind me fifteen minutes beforehand \\because i want to be prepared for the meeting} & 0.9151366 & (ReceivesAction, UsedFor) & (prepare for meet, prepare for) \\ \hline
\thead[l]{if we are approaching fall \\then remind me to buy flower bulbs \\because i want to make sure i have a pretty spring garden} & 1.0 & (CausesDesire, CausesDesire) & (plant flower, plant flower) \\ \hline
\thead[l]{if i receive emails about sales on basketball shoes \\then let me know about the sale \\because i want to save money} & 0.9345853 & (CausesDesire, UsedFor) & (buy something, i buy something) \\ \hline
\thead[l]{if i receive an email related to work \\then notify me about the email immediately \\because i want to stay on top of my work-related emails} & 1.0 & (CausesDesire, CausesDesire) & (open email, open email) \\ \hline
\thead[l]{if the forecast is dry and greater than 50 degrees f on a weekend day \\ then remind me to line-dry the laundry \\ because i want our clothes to smell good} & 1.0 & (MotivatedByGoal, CapableOf) & (smell good, smell good) \\ \hline
\thead[l]{if i have more than three hours meeting on my calendar for a day \\ then remind me to relax for an hour in the evening \\ because i want to achieve work life balance} & 0.96545255 & (CapableOf, CapableOf) & (make me feel good, make me look good) \\ \hline
\end{tabular}
}
\end{table*}

Tables \ref{tab:half_proof_ex_p1} and \ref{tab:half_proof_ex_p2} show 32 examples of the 67 expert-verified half-proofs reported in Table \ref{tab:n_hop}. These half-proofs are obtained using bidirectional 2-hop beam search and logically prove the \textGoal $\coloneq$ \textAction implication. Recall that \textAction refers to the Then-portion of the command and \textGoal refers to the Because-portion of the command. The embeddings used to measure semantic closeness in the prover is GloVe. The second intermediate hop obtained by \coolname's prover is shown in column 4 of the tables. Each row indicates a successfully half-proved if-then-because command along with its intermediate commonsense reasoning hop. Let us explain the proofs through an example. Consider the example on Row 5 of Table \ref{tab:half_proof_ex_p1}: 
``if the temperature is going to be below 40 degrees in the evening but above 40 degrees in the morning then remind me to bring a jacket because i want to stay warm on my commute''. The relations (column 3) and \comet{} outputs (column 4) for this example indicate that the \textAction of reminding the user to bring a jacket \emph{causesDesire} for the user to \emph{wear the jacket} which is a \emph{Prerequisite} for the \textGoal of staying warm on the user's commute. In other words \coolname{} is able to understand why reminding someone to bring a jacket would allow them to stay warm (because they would be able to wear the jacket). This is a logical proof for the implication \textGoal $\coloneq$ \textAction~according to the proof definition of the prover proposed in this paper. For more examples, please refer to Tables \ref{tab:half_proof_ex_p1} and \ref{tab:half_proof_ex_p2}.

\end{document}